\documentclass{amia}
\usepackage{graphicx}
\usepackage[labelfont=bf]{caption}
\usepackage[superscript,nomove]{cite}
\usepackage{color}
\usepackage{xcolor}
\usepackage{hyperref}
\usepackage{subcaption}
\usepackage{booktabs}
\usepackage{multirow}
\usepackage{changepage}
\usepackage{colortbl}
\usepackage{amsmath}
\usepackage{amssymb}
\usepackage{marginnote}
\usepackage{multicol}

\begin{document}

\newcommand{\todo}[2][NA]{{\color{red} $\rhd$\marginnote{[Assignee #1]: #2}}}
\newcommand{\note}[2][NA]{{\reversemarginpar \color{blue} $\lhd$\marginnote{[#1 says]: #2}}}

\title{Blending Knowledge in Deep Recurrent Networks for Adverse Event Prediction at Hospital Discharge}

\author{Prithwish Chakraborty, PhD$^{1}$;
        James Codella, PhD$^{1}$;
        Piyush Madan, MS$^{1}$;
        Ying Li, PhD$^{1}$;
        Hu Huang, PhD$^{2}$;
        Yoonyoung Park, PhD$^{1}$;
        Chao Yan, MS$^{3}$;
        Ziqi Zhang, BSc$^{3}$;
        Cheng Gao, PhD$^{4}$;
        Steve Nyemba, MS$^{4}$;
        Xu Min, PhD$^{1}$;
        Sanjib Basak, MS$^{2}$;
        Mohamed Ghalwash, PhD$^{1}$;
        Zach Shahn, PhD$^{1}$;
        Parthasararathy Suryanarayanan, BSc B.Tech$^{1}$;
        Italo Buleje, MS$^{1}$;\\
        Shannon Harrer, PhD$^{2}$;
        Sarah Miller, PhD$^{1}$;
        Amol Rajmane, MS$^{2}$;
        Colin Walsh, MD MA$^{3}$;\\
        Jonathan Wanderer, MD MPhil$^{3}$;
        Gigi Yuen Reed, PhD$^{2}$;\\
        Kenney Ng, PhD$^{2}$;
        Daby Sow, PhD$^{1}$;
        Bradley A. Malin, PhD$^{3,4}$}

\institutes{
    $^1$IBM Research, USA;
    $^2$IBM Watson Health, USA;\\
    $^3$Vanderbilt University, Nashville, TN, USA;\\
    $^4$Vanderbilt University Medical Center, Nashville, TN, USA\\
}

\maketitle

\noindent{\bf Abstract}

\textit{Deep learning architectures
have an extremely high-capacity for modeling complex
data in a wide variety of domains.  However, 
these architectures have been limited in their ability to support complex prediction problems using insurance claims data, such as readmission at 30 days, mainly due to data sparsity issue. Consequently, classical machine learning methods, especially those that embed domain knowledge in handcrafted features, are often on par with, and sometimes  outperform, deep learning approaches. 
In this paper, we illustrate how the potential of deep learning can be achieved by blending domain knowledge within deep learning architectures
to predict adverse events at hospital discharge, including readmissions. More specifically, we introduce a learning architecture that fuses a representation of patient data computed by a self-attention based recurrent neural network, with clinically relevant features. We conduct extensive experiments on a large claims dataset and show that the blended method outperforms the standard
machine learning approaches. }

\section*{Introduction}
    The digitization of health data
has sparked artificial intelligence (AI) researchers to develop novel computational methods 
for various 
tasks, including the sequential prediction of clinically meaningful events, such as hospital readmissions and death. 
Initially, these methods were based on classical machine learning techniques, ranging from simple
parametric regression
to more complex non-parametric models, such as decision
trees and rule learning techniques.
Yet classical models are limited in their representative capacity and make certain assumptions about the data distribution (e.g., linearity assumptions) that do not always hold true.  As such, deep learning architectures, including convolutional neural networks,  
recurrent neural networks,
and, more recently, attention schemes have illustrated better performance potential for a variety of clinical prediction problems. We refer the reader to a recent review by Cao and colleagues \cite{Cao:JAMIA2018} for a comprehensive survey on this topic.

With respect to the readmission prediction problem, there have been several studies illustrating the potential for deep learning. 
First, a 30-day readmission prediction for patients with congestive heart failure (CHF) was achieved through the novel TopicRNN architecture, which combines global and local context \cite{xiao2018readmission}. Though the performance was modest  (AUC in the $0.60 - 0.65$ range ), this architecture outperformed
state-of-the-art recurrent neural architectures, including RETAIN algorithm \cite{RETAIN}. 
More recently, predictive performance was improved to over 0.70 AUC for CHF patients through a cost-sensitive formulation of a long short-term memory model that incorporates expert features and contextual embedding of clinical concepts\cite{ashfaq2019readmission}.

Despite its potential,
deep learning has been limited in its adoption by the healthcare community and specifically when applied to electronic health records (EHR) and claims data. As summarized by Wang and colleagues \cite{wang2019deep}, while deep learning architectures have
achieved success in the medical imaging 
domain \cite{gulshan2016,esteva2017}, it has been more challenging to translate into
solutions for EHR and claims data. The lack of translation stems from a number of factors, including those instigated by the data itself, such as high feature heterogeneity (e.g., a combination of discrete, continuous, and categorical features) and variability in quality, and 
those instigated by modeling (e.g., interpretability of learned features and their weights). Moreover, many healthcare organizations are wary of the generalizability of the resulting models, a concern driven by challenges in semantic interoperability when these models are applied beyond the context in which they are initially trained. As an illustration of the challenge, it was recently shown that deep learning approaches were unable to outperform conventional methods in hospital
readmission prediction \cite{xumin:NSR2019}. More specifically, when only insurance claims data is available, Lasso regression appears to achieve readmission prediction performance that is comparable to deep neural network architectures \cite{goto2019,Allam2019NeuralNV}.

 As a consequence,
though classical machine models may be suboptimal to their deep learning descendants, 
the former
can perform well given the limited and sparse nature
of medical data — especially when significant effort is applied to 
handcraft the features that represent
domain-specific knowledge. These classical approaches also benefit from better alignment with current medical knowledge and, thus, are more readily
interpretable to end users. Based on these observations, the AI community leadership has warned the community about the negative consequences that could transpire from overuse of deep approaches that tend to be less generalizable while being more difficult to interpret when more conventional rule learning methods could be applied \cite{rudin2019stop}.

In this paper, we illustrate that blending knowledge into deep learning frameworks can improve adverse event prediction at hospital discharge, including readmission and mortality. Specifically, we introduce a prediction method for such events from claims data that integrates 1) deep recurrent models, which capture complex temporal partners in patient data,  with 2) knowledge-driven approaches, which capture medically relevant aspects of patient states.
We demonstrate experimentally 
the benefits of this symbiosis between these data driven and knowledge driven modeling approaches. We applied this approach on a large medical claims data set, thus providing models that can be used by health organizations wishing to improve quality scores and reduce potential revenue losses resulting from these adverse events.

\begin{figure}[h]
    \centering
    \begin{subfigure}[b]{0.49\textwidth}
    \begin{center}
      \includegraphics[width=\linewidth]{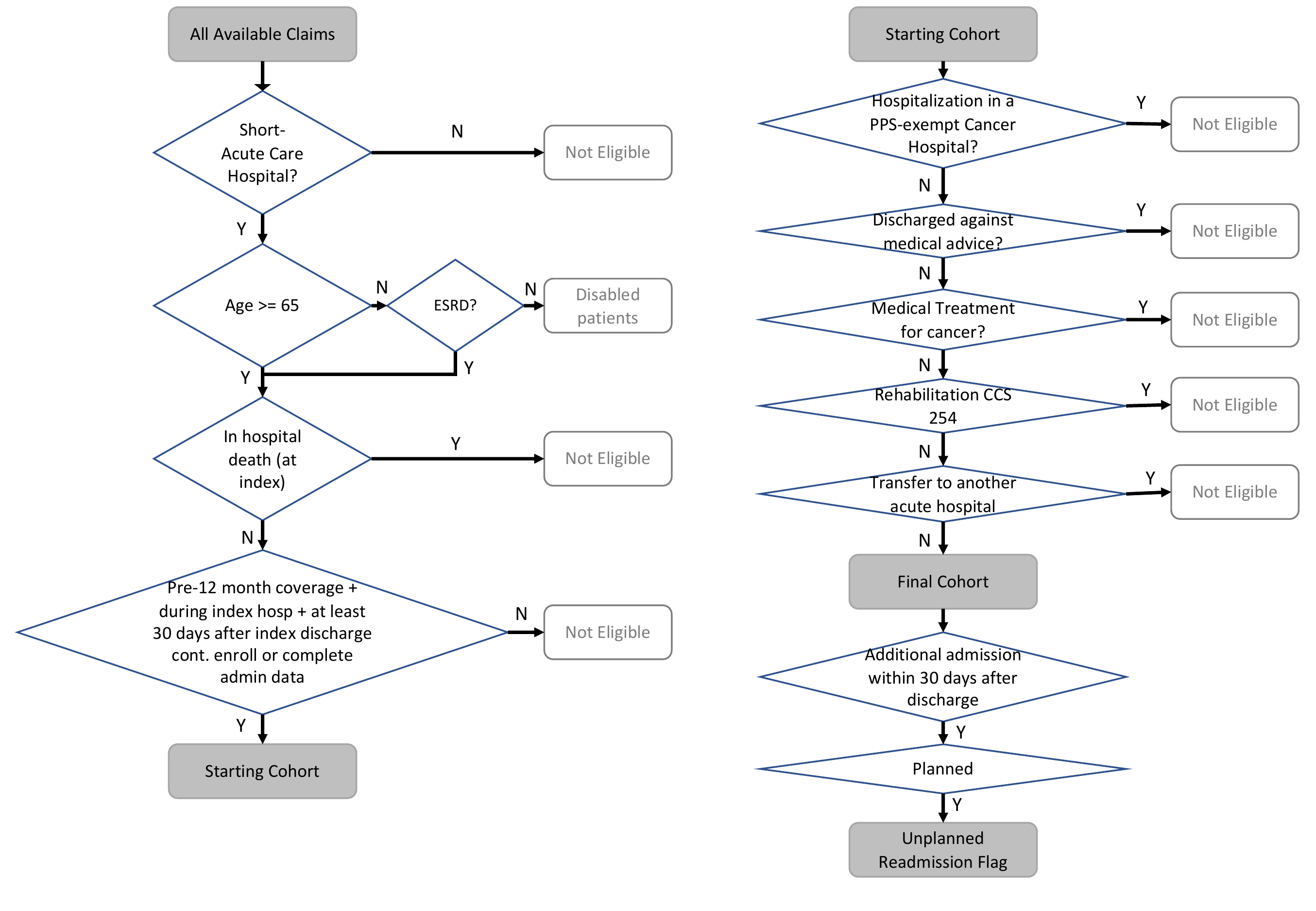}
    \end{center}
      \caption{Cohorts construction steps}
      \label{fig:cohortFlow}
    \end{subfigure}
\begin{subfigure}[b]{0.5\textwidth}
      \begin{center}
      \includegraphics[width=\linewidth]{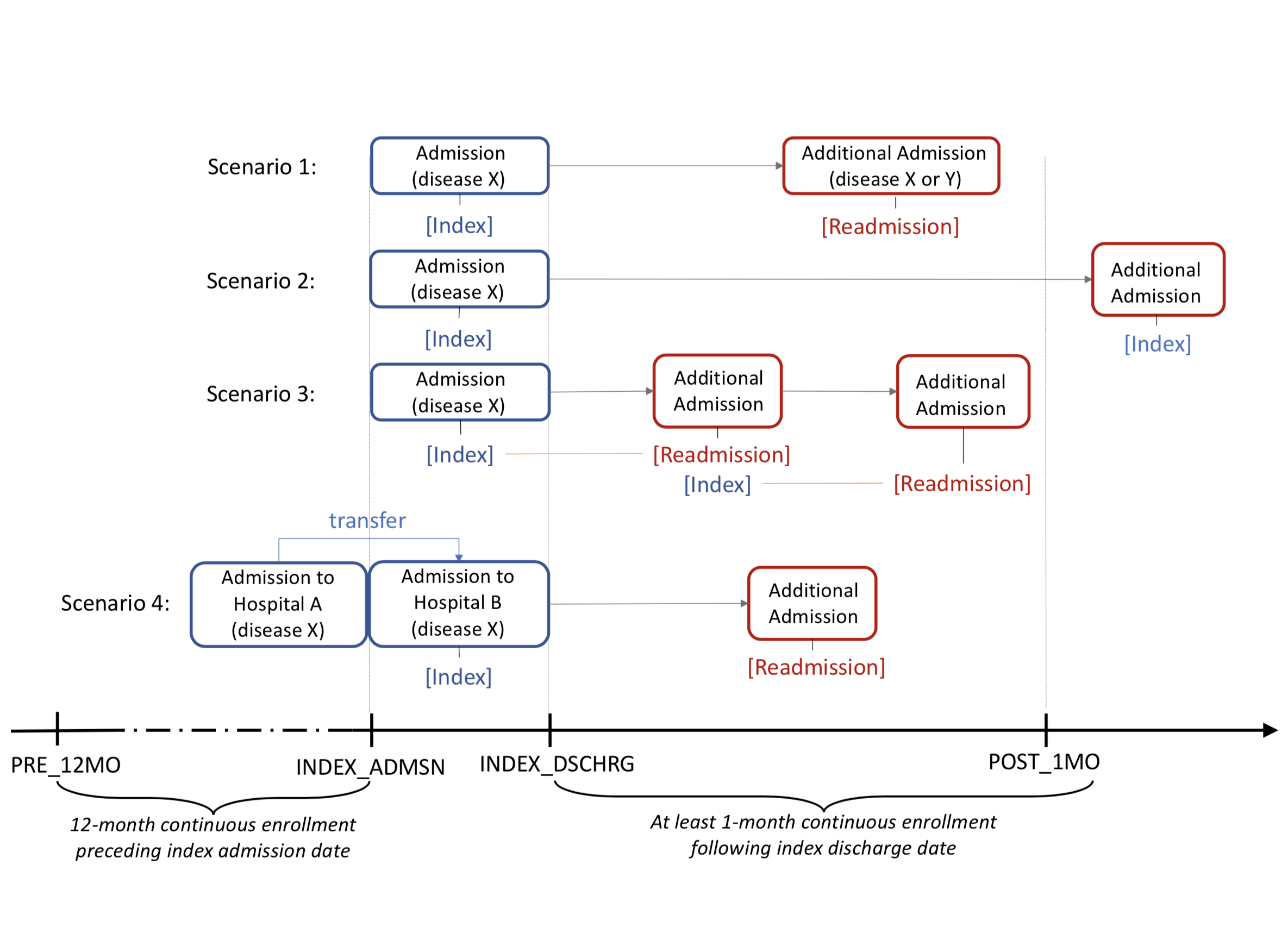} 
      \end{center}
      \caption{Scenarios for target events} 
      \label{fig:cohortScenarios}
    \end{subfigure}
    \caption{Cohort construction description: (a) shows the detailed steps for creating the index and target events and (b) shows the various scenarios to consider while resolving consecutive claim periods towards an index event.}
    \label{fig:cohortFull}
\end{figure}
\section*{Methods}

\begin{table}[tbh]
  \centering
  \caption{Descriptive statistics for the cohorts assembled for this study.\label{tab:counts}}
  \scriptsize
  \begin{adjustwidth}{0.2cm}{}
    \begin{tabular}{|l|l|l|l|l|l|l|l|l|l|}
\hline
\multirow{9}{*}{\begin{tabular}[c]{@{}l@{}}Final Cohort (2011)               \\ Total beneficiaries: 220,093\end{tabular}}                      & \cellcolor{gray!80}RACE       & \cellcolor{gray!80}Unknown & \cellcolor{gray!80}White         & \cellcolor{gray!80}Black      & \cellcolor{gray!80}Other      & \cellcolor{gray!80}Asian      & \cellcolor{gray!80}Hispanic   & \multicolumn{1}{c|}{\begin{tabular}[c]{@{}c@{}}\cellcolor{gray!80}North \\ \cellcolor{gray!80}American \\ \cellcolor{gray!80}Native\end{tabular}} & \cellcolor{gray!80}Total  \\ \cline{2-10} 
                                                                                                                                                & Counts     & 255     & 190,767       & 18,945     & 2,189      & 2,873      & 4,086      & 978                                                                                      & 220,093 \\ \cline{2-10} 
                                                                                                                                                & Percentage & 0.12\%  & 86.68\%       & 8.61\%     & 0.99\%     & 1.31\%     & 1.86\%     & 0.44\%                                                                                   &        \\ \cline{2-10} 
                                                                                                                                                & \cellcolor{gray!50}Gender     & \cellcolor{gray!50}Male    & \cellcolor{gray!50}Female        &        \cellcolor{gray!50}    &         \cellcolor{gray!50}   &        \cellcolor{gray!50}    &        \cellcolor{gray!50}    &   \cellcolor{gray!50}                                                                                      & \cellcolor{gray!50}Total  \\ \cline{2-10} 
                                                                                                                                                & Counts     & 91,389  & 128,704       &            &            &            &            &                                                                                          & 220,093 \\ \cline{2-10} 
                                                                                                                                                & Percentage & 41.52\% & 58.48\%       &            &            &            &            &                                                                                          &        \\ \cline{2-10} 
                                                                                                                                                & \cellcolor{gray!40}Age Range  & \cellcolor{gray!40}Unknown & \cellcolor{gray!40}\textless{}65 & \cellcolor{gray!40}65$\sim$69 & \cellcolor{gray!40}70$\sim$74 & \cellcolor{gray!40}75$\sim$79 & \cellcolor{gray!40}80$\sim$84 & \cellcolor{gray!40}\textgreater{}85                                                                         & \cellcolor{gray!40}Total  \\ \cline{2-10} 
                                                                                                                                                & Counts     & 7,602   & 0             & 39,843     & 43,374     & 42,735     & 42,154     & 44,385                                                                                   & 220,093 \\ \cline{2-10} 
                                                                                                                                                & Percentage & 3.45\%  & 0.00\%        & 18.10\%    & 19.71\%    & 19.42\%    & 19.15\%    & 20.17\%                                                                                  &        \\ \hline
\multicolumn{10}{|l|}{}                                                                                                                                                                                                                                                                                                                        \\ \hline
\multirow{9}{*}{\begin{tabular}[c]{@{}l@{}}30-day Readmission (2011)                 \\ Total beneficiaries: 33,236\end{tabular}}               & \cellcolor{gray!80}RACE       & \cellcolor{gray!80}Unknown & \cellcolor{gray!80}White         & \cellcolor{gray!80}Black      & \cellcolor{gray!80}Other      & \cellcolor{gray!80}Asian      & \cellcolor{gray!80}Hispanic   & \multicolumn{1}{c|}{\begin{tabular}[c]{@{}c@{}}\cellcolor{gray!80}North \\ \cellcolor{gray!80}American \\ \cellcolor{gray!80}Native\end{tabular}} & \cellcolor{gray!80}Total  \\ \cline{2-10} 
                                                                                                                                                & Counts     & 47      & 28,087        & 3,504      & 321        & 408        & 710        & 159                                                                                      & 33,236  \\ \cline{2-10} 
                                                                                                                                                & Percentage & 0.14\%  & 84.51\%       & 10.54\%    & 0.97\%     & 1.23\%     & 2.14\%     & 0.48\%                                                                                   &        \\ \cline{2-10} 
                                                                                                                                                & \cellcolor{gray!50}Gender     & \cellcolor{gray!50}Male    & \cellcolor{gray!50}Female        &     \cellcolor{gray!50}       &         \cellcolor{gray!50}   &       \cellcolor{gray!50}     &      \cellcolor{gray!50}      &                                                                                      \cellcolor{gray!50}    & \cellcolor{gray!50}Total  \\ \cline{2-10} 
                                                                                                                                                & Counts     & 14,225  & 19,011        &            &            &            &            &                                                                                          & 33,236  \\ \cline{2-10} 
                                                                                                                                                & Percentage & 42.80\% & 57.20\%       &            &            &            &            &                                                                                          &        \\ \cline{2-10} 
                                                                                                                                                & \cellcolor{gray!40}Age Range  & \cellcolor{gray!40}Unknown & \cellcolor{gray!40}\textless{}65 & \cellcolor{gray!40}65$\sim$69 & \cellcolor{gray!40}70$\sim$74 & \cellcolor{gray!40}75$\sim$79 & \cellcolor{gray!40}80$\sim$84 & \cellcolor{gray!40}\textgreater{}85                                                                         & \cellcolor{gray!40}Total  \\ \cline{2-10} 
                                                                                                                                                & Counts     & 1,268   & 0             & 5,390      & 6,253      & 6,520      & 6,764      & 7,041                                                                                    & 33,236  \\ \cline{2-10} 
                                                                                                                                                & Percentage & 3.82\%  & 0.00\%        & 16.22\%    & 18.81\%    & 19.62\%    & 20.35\%    & 21.18\%                                                                                  &        \\ \hline
\multicolumn{10}{|l|}{}                                                                                                                                                                                                                                                                                                                        \\ \hline
\multirow{9}{*}{\begin{tabular}[c]{@{}l@{}}Unplanned 30-day \\ readmission (2011)                  \\ Total beneficiaries: 18,329\end{tabular}} & \cellcolor{gray!80}RACE       & \cellcolor{gray!80}Unknown & \cellcolor{gray!80}White         & \cellcolor{gray!80}Black      & \cellcolor{gray!80}Other      & \cellcolor{gray!80}Asian      & \cellcolor{gray!80}Hispanic   & \multicolumn{1}{c|}{\begin{tabular}[c]{@{}c@{}}\cellcolor{gray!80}North \\ \cellcolor{gray!80}American \\ \cellcolor{gray!80}Native\end{tabular}} & \cellcolor{gray!80}Total  \\ \cline{2-10} 
                                                                                                                                                & Counts     & 28      & 15,200        & 2,148      & 190        & 261        & 416        & 86                                                                                       & 18,329  \\ \cline{2-10} 
                                                                                                                                                & Percentage & 0.15\%  & 82.93\%       & 11.72\%    & 1.04\%     & 1.42\%     & 2.27\%     & 0.47\%                                                                                   &        \\ \cline{2-10} 
                                                                                                                                                & \cellcolor{gray!50}Gender     & \cellcolor{gray!50}Male    & \cellcolor{gray!50}Female        &         \cellcolor{gray!50}   &        \cellcolor{gray!50}    &           \cellcolor{gray!50} &      \cellcolor{gray!50}      &                                                                    \cellcolor{gray!50}                      & \cellcolor{gray!50}Total  \\ \cline{2-10} 
                                                                                                                                                & Counts     & 7,976   & 10,353        &            &            &            &            &                                                                                          & 18,329  \\ \cline{2-10} 
                                                                                                                                                & Percentage & 43.52\% & 56.48\%       &            &            &            &            &                                                                                          &        \\ \cline{2-10} 
                                                                                                                                                & \cellcolor{gray!40}Age Range  & \cellcolor{gray!40}Unknown  \cellcolor{gray!40}\textless{}65 & \cellcolor{gray!40}65$\sim$69 & \cellcolor{gray!40}70$\sim$74 & \cellcolor{gray!40}75$\sim$79 & \cellcolor{gray!40}80$\sim$84 & \cellcolor{gray!40}\textgreater{}85                                                                         & \cellcolor{gray!40}Total  \\ \cline{2-10} 
                                                                                                                                                & Counts     & 681     & 0             & 3,220      & 3,639      & 3,643      & 3,724      & 3,422                                                                                    & 18,329  \\ \cline{2-10} 
                                                                                                                                                & Percentage & 3.72\%  & 0.00\%        & 17.57\%    & 19.85\%    & 19.88\%    & 20.32\%    & 18.67\%                                                                                  &        \\ \hline
\end{tabular}

   \end{adjustwidth}
\end{table}

\subsection*{Data and Study Setting}
In this section, we start by describing our study setting. We primarily focused on prediction of unplanned readmission and mortality 30 days from discharge from a large scale claims dataset spanning millions of patients. 
Unplanned readmissions are undesirable events that can lead to increased healthcare costs and poorer health outcomes for patients. 
For this study, from the claims dataset we constructed the cohort by extracting the selected patients and identifying the prediction target events for 30-day unplanned readmission and also 30-day post discharge mortality.
To do so, we first identify and define an \emph{index event} from which we predict risk of each 
outcome as the discharge date from an inpatient admission.  
We begin with all claims for patients continuously enrolled for at least 12 months prior to an index event and for at least 30 days following discharge.
Claims identified as index events must satisfy the following criteria:

$\bullet$ admissions must be for short stays requiring acute care. 
\\ $\bullet$ age of patient (except those who are eligible based due to end stage renal disease) at admission must be $\geq$ 65. 
\\ $\bullet$  patient must not expire during the inpatient stay. Note that these claims are not considered as index events but can be used to predict unexpected mortality if the previous index event was within 30 days of the death.
\\ $\bullet$ patient must not have been transferred to another acute care hospital at discharge – in this case we examine the last claim in a series of one or more transfers as a possible index event.

These criteria are mainly driven by the guidelines outlined by  Horowitz and colleagues. \cite{horowitz1,horowitz2} for general cohort selection.
After identifying the index events, we extract eligible \emph{target events} (unplanned readmissions and mortality) within 30 days of 
each index event. 
Figure~\ref{fig:cohortFlow} illustrates the cohort construction process. 
Another important step pertains to resolving continuous periods of claims as a single claim. Figure~\ref{fig:cohortScenarios} illustrates various scenarios that can arise when matching index events with readmissions. 
Scenario 1 shows a readmission that satisfies the 30-day readmission criteria whereas scenario 2, depicts a readmission 
outside the 30-day period and thus not considered to be a target event. 
Furthermore, we define a readmission in reference to an initial inpatient claim, such that only the first readmission within 30 days is considered as a target event. Scenario 3 shows a sequence of admissions where the second admission is considered to be a readmission for the first admission. The third admission is considered to be a readmission 
for the second admission only, in spite of occurring within 30 days of the first admission. It is to be noted, that all admissions including readmissions can qualify as index events if these meet the criteria for index event selection described earlier. 
Scenario 4 shows how we combine the transfers as a single event. 

{\noindent\bf Defining the Target events:}
To define unplanned readmission, we first identified admissions that were considered planned. 
Any readmission, that was not planned or an acute event, was designated as unplanned. Planned readmissions are defined using a pre-specified list of 
procedure codes, or diagnostic codes for maintenance chemotherapy or rehabilitation. The list of codes were based on existing literature 
(See \cite{horowitz2} Table 1 and Table 2). Even if planned procedures occurred, readmissions for acute illness or for complications 
of care were not considered planned. The principal diagnostic code was used to identify such admissions. 
As a secondary target to evaluate our models, we defined unexpected 30-day mortality target. 
To find \emph{unexpected mortality}, from each index claims, we identified patients who were dead within 30 days after discharge from an index admission based on the denominator table. We excluded patients who left the hospital against medical advice and died expired within 30 days of those discharge. We also excluded patients who expired after getting admitted to hospice. While the first criteria is standard and used by agencies such as the Centers for Medicare and Medicaid Services (CMS), the second criteria was included as patients in hospice care follows quite different patterns of care. 
Counts for the selected events are shown in Table~\ref{tab:counts}.

\subsection*{Model Description}

Claims data are typically multi-modal. Such data sets are also very high-dimensional and sparse. In addition, the historical data for each index admission is of varying temporal length. Our data set covers diagnosis, procedures, and demographic information. 

In practice, methods such as LACE scores have been used to score the probability of unplanned readmissions.
The LACE index~\cite{van2010derivation} is computed using four dimensions to predict the risk of mortality 
or non-elective readmission 30 days after discharge. These variables are: 
length of stay (L), acuity of the admission (A), comorbidity of the patient (C), and emergency department visits (E) in the past 6 months from the index event. Using these, a 19 point scale is derived which can then be used to define the risk. While LACE provides a very interpretable and easy-to-use method to compute risks, it doesn't take into account the temporal information of patients from many different data sources. 

\begin{figure}[htpb]
  \begin{center}
  \includegraphics[width=0.9\linewidth]{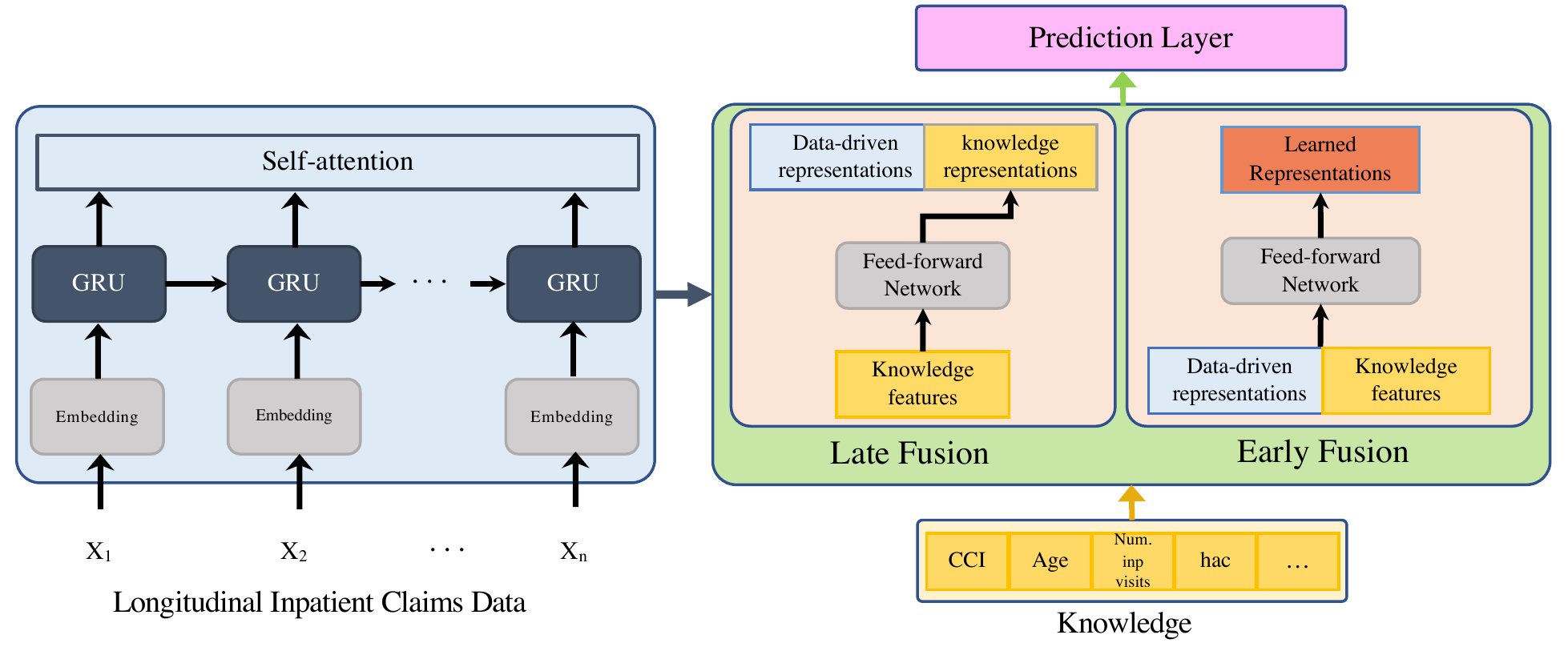}
\end{center}

  \caption{\textit{Blending domain knowledge into Deep Recurrent Neural Networks}. Left (blue) box shows a self-attention based RNN that processes longitudinal inpatient claims data. The sparse data is embedded to a lower compact dimension using a Linear Embedding or Med2Vec layer. Healthcare domain-specific features (yellow) are next ingested into the model using either (a) early fusion or (b) late fusion strategy (Green box, right). 
  In early fusion, domain-specific features are concatenated with representations from the RNN and the final model output is obtained using a
  deep feed-forward network.
  In late fusion, the domain features are transformed using a deep feed-forward network whose output is concatenated
  with representations from the RNN to a single shallow layer to generate model outputs.}\label{fig:model_gru}
\end{figure}

\textbf{RNNs to model longitudinal claims data}: 
To effectively model the longitudinal claims data, we built a Recurrent Neural Network (RNN) model that ingests the temporal history of the patients and produces a risk score for the relevant target as its prediction. 
The architecture of the RNN is shown on the left side of Figure~\ref{fig:model_gru} (blue box) which we term as the data-driven model. The following is a list of three salient aspects of this RNN data-driven model.
\\ $\bullet$ To account for the high dimensionality of patient records, we group together diagnosis and procedures using Clinical Classification Software (CCS) categories at their lowest level. Using this transformation we are left with more than 400 features at every time point which are one-hot encoded.
\\ $\bullet$ Our dataset is highly sparse and high dimensional - to effectively model this dataset we apply an embedding layer to transform the raw one-hot encoded features at each time point to a compact representation. 
\\ $\bullet$ Finally, we feed the embedded vector to an RNN with self-attention to process the longitudinal temporal data.

There has been significant amount of work in literature in finding effective embedding strategies that are applicable to health data. Med2Vec~\cite{med2vec}
is one such method that is
generally applicable. It fits a shallow network that aims to discover vector representations that are consistent between visits, while accounting for the apparent lack of order between features in a single visit. As an alternative mechanism we also fit a 1-layer feed-forward network to produce compact representations of the input data.
While we experimented with different RNN architectures, we settled on a Gated Recurrent Unit (GRU) as our base RNN based on its performance. While such models have been successfully applied to computational health modeling scenarios~\cite{RETAIN,Cao:JAMIA2018}, it has recently been observed that adding an attention layer can significantly improve the performance of such deep networks.

Formally, let us denote the medical history of the $n^{th}$ patient (where  $n\in{1,\dots,N}$) by $X^n = \lbrace X^n_1, X^n_2, \cdots, X^n_{T^n}\rbrace$, 
where $T^n$ is the total number of observed time points for patient $n$ and  $X^n_t \in \mathcal{R}^M$ represents the 
observed features. 
For the sake of simplicity, we drop the superscript $n$ in the rest of this paper.
Input data $X_t$ at each time point $t$ is ingested by this RNN architecture (see Figure~\ref{fig:model_gru}), starting with several layers, to produce an embedding vector $e_t$ using either Med2Vec or a feed-forward network as $e_t = W_e e_t + b_e$. This information is combined with information from historical data points represented as a hidden vector $h_t$ as $h_{t+1} = \text{GRU}(e_t, h_t)$.
Finally, we use self-attention\cite{selfatt} to combine all the hidden states $H=\lbrace h_0, h_1, ..., h_T\rbrace$ and produce the final output $y_t$ as the prediction for the relevant problem (readmission/mortality) as:
\vspace{-3em}
\begin{multicols}{2}
  \null \vfill
  \begin{equation}
  a_t  = \text{Softmax}(\frac{h_TH}{\sqrt{|h_T|}})H \text{;}\qquad o_t = a_t \boldsymbol{\cdot} H \label{eq:attention}
  \end{equation}
  \vfill \null
  \columnbreak  \null \vfill
  \begin{equation}
    y_t  = \text{Sigmoid}\left(W_o o_T + b_o \right)
     \label{eq:output}
  \end{equation}
  \vfill \null
\end{multicols}
\vspace{-2em}

We performed experiments that included, as well as neglected, outpatient claims.  We observed that such claims had minimal influence on  model performance and were excluded from further investigation.

\textbf{Availability of domain knowledge}: 
Deep learning models, such as the one described above, are non-parametric and have 
a high capacity to model various complicated concepts in observational data. 
However, in limited and sparse data settings, such as that encountered  in medical data, classical methods can perform well - especially when significant effort has been applied to handcraft the features. 
These features are quite beneficial to end users consuming the outputs of such predictions as they are readily interpretable and transparent. 
In fact, it was recently reported that deep learning approaches are unable to outperform such conventional methods to predict hospital readmissions\cite{xumin:NSR2019}. 
It should be noted that in well studied problems, such as readmission prediction, there is a
significant literature about how to construct such features~\cite{horowitz1,horowitz2}.
As such, we hypothesize  that we can improve the predictive performance of deep learning models - even under such data sparsity situations - by blending such features into deep learning models. 
To this effect, we constructed several features from domain knowledge as listed in Table~\ref{tab:features}.
These features were identified from literature and verified by subject matter experts.
They are categorized along four groups: (a) Comorbidity,
(b) Clinical, (c) Demographic, and (d) Others. The latter category tracks facility and
discharge disposition information that we used to derive discharge related
actionable insights. A special class of the features constructed from 
Clinical data elements are \emph{hospital acquired conditions} (HAC) or \emph{presence of hospital acquired complications during
index admission}.  HAC are undesirable events and can be indicative of future complications. The full list of HAC is shown in Table~\ref{tab:HAC} which we constructed by leveraging the HAC  as published by CMS.

\begin{table}[h!]
  \centering
  \caption{Features used: List of knowledge driven features (left) list of hospital acquired conditions (right)}
  \label{tab:all_feat}
  \begin{subtable}[t]{0.60\linewidth}
    \begin{center}
      \caption{Knowledge derived Features}
      \scriptsize
      \label{tab:features}
      \begin{tabular}{l|p{0.8\linewidth}}
    \toprule
    \textbf{Category} & \textbf{Definition}\\
    \midrule
    \multirow{2}{*}{Comorbidity} & DX \& PROC Categorization\\ 
    & Charlson Comorbidity Index \\
    \midrule
    \multirow{13}{*}{Clinical} & Length of Stay (index admission) \\ 
    & Number of Inpatient Admissions during Previous 12 Months \\ 
    & Number of Outpatient Visits during Previous 12 Months \\
    & Number of ED Visits during Previous 12 Months \\
    & Type of Index Admission \\
& Admission Source \\
    & Discharge Disposition \\
    & Discharge Diagnosis \\
    & List of Hospital Acquired Complications During Index Admission \\
    & Diagnosis Related Group \\
    & Number of DX Codes on a Claim \\
    \midrule
    \multirow{5}{*}{Demographic} & Age Group \\ 
    & Gender \\ 
    & Race Codes \\
& Dual Eligibility\\
    & Reason for Medicare Eligibility\\
    \midrule
    \multirow{1}{*}{Others} & Facility ID \\ 
\bottomrule
  \end{tabular}

     \end{center}
  \end{subtable}
\begin{subtable}[t]{0.35\linewidth}
    \begin{center}
      \caption{Hospital acquired conditions}
      \scriptsize
      \label{tab:HAC}
      \begin{tabular}{>{\raggedright\arraybackslash}p{0.66\linewidth}}
  \toprule
  \textbf{Hospital Acquired Complications} \\
  \midrule
  Foreign Object Retained After Surgery \\
  Air Embolism \\
  Blood Incompatibility \\
  Pressure Ulcer Stages III \& IV \\
  Falls and Trauma \\ 
  Catheter-Associated Urinary Tract Infection (UTI) \\
  Vascular Catheter-Associated Infection \\
  Manifestations of Poor Glycemic Control \\
  Surgical Site Infection, Mediastinitis, Following Coronary Artery Bypass Graft (CABG) \\
  Surgical Site Infection Following Certain Orthopedic Procedures \\
  Surgical Site Infection Following Bariatric Surgery for Obesity \\
  Deep Vein Thrombosis and Pulmonary Embolism Following Certain Orthopedic Procedure \\
  \bottomrule
\end{tabular}
     \end{center}
  \end{subtable}
\end{table}

\textbf{Blending domain knowledge in an RNN}:
To effectively use such domain knowledge directly in our deep learning architecture
we fuse this knowledge with the RNN output as shown on the right side of Figure~\ref{fig:model_gru} in the green box. We refer to this as the \emph{Fusion} layer and design two strategies to blend this information: (a) Early fusion and (b) Late fusion. 
In early fusion, we concatenate the domain features $z$ and learned data representations $o_T$ from Equation~\ref{eq:attention} and feed into a multi-layered feed forward network. By contrast, in late fusion we learn a representation of the domain knowledge by feeding it into a multi-layered feed-forward network and concatenate the data and knowledge representations. Finally, the output from each respective strategy is fed into an output prediction layers (Figure~\ref{fig:model_gru}, pink box) as described in equation~\ref{eq:output}. Formally this can be defined as follows:
\begin{equation}
 \begin{array}{rlclrlcl}
     \text{\bf Early Fusion:} & \hat{o}_T & = &\text{MLP}([o_T;z]) 
     & \qquad
     \text{\bf Late Fusion:} & \hat{o}_T & = & [o_T;\text{MLP}(z)] \\
 \end{array}
\end{equation}

\textbf{Baselines}:
We evaluated our models against 30-day readmission prediction and, another closely 
related task, 30-day unexpected mortality. We split the dataset, stratified by number of events (e.g., readmission episodes) for a patient, for each of the
problems randomly into training ($70\%$), validation ($15\%$), calibration
($5\%$), and test folds ($10\%$). 
The stratification ensured that the complete
history of a patient belonged to only one data subset. 
We evaluated both classical models (Logistic Regression, Random Forest, and Histogram-based Gradient Boosting Tree) as well as our proposed methods, henceforth referred to as `Early' and `Late' fusion. 
As the dataset is highly sparse, we compared models under both linear and med2vec embedding strategies. 
The tasks are also highly unbalanced (See Table~\ref{tab:counts}) such that failing to predict a true positive event may have a higher burden than over-predicting. To account for this aspect, we used state-of-the-art techniques such as SMOTE~\cite{chawla2002smote}for classical models. For deep models, we used a weighted loss between positives and negatives allowing us to penalize false negatives higher than false positives. While this is a standard setting for unbalanced data problems, for clinical tasks such as readmission predictions the final score from a predictive model is usually desirable to correlate with the probability of the event (so that these can be used as risk scores). Thus, once the models were trained we applied \emph{Platt Scaling} for classical models and
\emph{temperature scaling} for deep learning models on the calibration set to calibrate the score estimates from the models - this ensures that these scores have a probabilistic interpretation and $0.5$ can be chosen as the threshold to declare presence/absence. Our choices were
motivated by existing literature~\cite{guo2017calibration} that 
found these choices to work well in practice.

\section*{Results}

\begin{table}[t!] 
\begin{center}
    \caption{Comparison of model performance (w/ linear embedding vs.
    med2vec)
for prediction of two adverse events at hospital
    discharge (a) unplanned readmission and (b) unexpected mortality. Uncertainty based on top 10 models.}
\scriptsize
\begin{subtable}[t]{0.45\textwidth}
\caption{Readmission}
        \label{tab:readres}
    \begin{tabular}{lllllll}
  \toprule
  \textbf{Algorithm} & \multicolumn{2}{c}{With Linear Emb} & \multicolumn{2}{c}{With Med2Vec}\\
  \cmidrule{2-3} 
  \cmidrule{4-5}
                        & \textbf{AUC}  & \textbf{Recall} & \textbf{AUC} & \textbf{Recall}\\
  \midrule
  LR           & 0.628 ($\pm 0.004$) & 0.647 & 0.629 ($\pm 0.003$) & 0.554 \\
  RF           & 0.600 ($\pm 0.008$) & 0.712 & 0.478 ($\pm 0.001$) & 0.864 \\
  HistGBT      & 0.558 ($\pm 0.003$) & 0.218 & 0.442 ($\pm 0.002$) & 0.826 \\
  Early Fusion & 0.678 ($\pm 0.004$) & 0.549 & 0.654 ($\pm 0.005$) & 0.54\\
  Late Fusion  & 0.676 ($\pm 0.005$) & 0.51  & 0.652 ($\pm 0.006$) & 0.496 \\
  \bottomrule
\end{tabular}
     \end{subtable}
    \hspace{3em}
    \begin{subtable}[t]{0.45\textwidth}
\caption{Unexpected mortality}
\label{tab:mort}
    \begin{tabular}{lllllll}
  \toprule
  \textbf{Algorithm} & \multicolumn{2}{c}{With Linear Emb} & \multicolumn{2}{c}{With Med2Vec}\\
  \cmidrule{2-3} 
  \cmidrule{4-5}
                        & \textbf{AUC}  & \textbf{Recall} & \textbf{AUC} & \textbf{Recall} \\
  \midrule
  LR           & 0.800 ($\pm 0.006$) & 0.7839 & 0.774 ($\pm 0.004$) & 0.784 \\
  RF           & 0.785 ($\pm 0.004$) & 0.119  & 0.492 ($\pm 0.003$) & 0.697 \\
  HistGBT      & 0.766 ($\pm 0.003$) & 0.095  & 0.471 ($\pm 0.001$) & 0.557 \\
  Early Fusion & 0.840 ($\pm 0.008$) & 0.811  & 0.800 ($\pm 0.004$) & 0.687\\
  Late Fusion  & 0.844 ($\pm 0.005$) & 0.729  & 0.800 ($\pm 0.001$) & 0.711 \\
\bottomrule
\end{tabular}
   \end{subtable}
  \end{center}
\end{table}

The deep learning models were implemented using `pytorch', the classical models using `scikit-learn', and were run on a cluster containing eight NVIDIA V100 GPUs. For each model,
hyperparameter optimization was conducted using a grid search over a predefined grid and the
hyperparameters were selected based on the model with the best performance on validation sets. 
All models were able to chose from the same set of features and the grid for the hyperparameters were individually tuned for each model following best recommendations. For example, for our models we mainly chose between hidden size ($8 - 128$), number of layers ($1 - 3$), and batch size ($8 - 128$). For classical models, we also generated standard summary of temporal features such as mean and counts of features.
We report the comparison of model performance on the $10\%$ hold-out test fold for readmission prediction in Table~\ref{tab:readres}. We report both area under the receiver operating characteristic curve (AUC)  as a general metric and recall to evaluate the false-negative rate. Uncertainty around AUC based on the top 10 best fitted models is also reported. To ensure that model outputs can be mapped to probabilities, we always chose $0.5$ as the threshold to calculate recall. 
We conducted initial experiments on readmission prediction problem where we compared a RNN only model (without any fusion) to the baselines. Our results were similar to ~\cite{xumin:NSR2019} where RNN models failed to outperform the baselines. 
Overall, for 30-day readmission prediction, the proposed deep models performed at acceptable (or slightly better)
AUC values, ranging from $0.652$ to $0.682$, compared to reported numbers from 
literature~\cite{xumin:NSR2019, morgan2019assessment} on general readmission
(not necessarily unplanned). We also notice that there was a performance gain with the deep learning models over classical machine learning models.  For example, for the 30-day readmission prediction problem, the best performing deep learning
model \emph{Early Fusion} achieved an AUC of $0.682$ and recall of $0.549$ while
the best performing conventional machine learning model is logistic regression with an AUC of $0.632$ and recall of $0.647$. 
{\it It should be noted, that our proposed models were successfully tuned to achieve very high Recall-at-top-k (close to 1) at the expense of lower precision}.
We also observed that the deep learning models can be tuned more-readily to balance the  trade-off between these metrics dependent on the performance indicators of interest such as AUC and Recall. 
For this study, we opted for models above an acceptable AUC threshold (e.g., $0.6$ for 30-day readmission prediction) and higher recall to account for the rarity of the tasks.\\
As a secondary study, Table~\ref{tab:mort} reports on the same performance measures for 30-day unexpected mortality. The proposed models
perform relatively well for this task, with the best AUC varying between $0.806$ to $0.849$
and recall ranging from $0.697$  to $0.811$ for the best models. 
It can be seen the deep learning models perform the best for this task as well with \emph{Early Fusion} and 
\emph{Late Fusion} achieving the top results.

\section*{Discussion}

\textbf{What did we learn from model selection - Does med2vec help Early or Late Fusion?} 
We trained a suite of models and optimized for the hyperparameters to find the best
validated version of each model. We performed a \emph{quasi-ablation} study
by selectively including and excluding different features such as outpatient claims and domain knowledge. 
Based on our experiments, it was found that the outpatient historical claims did not  improve model performance when incorporating both inpatient claims and domain features.
Also, we found that in general, the use of med2vec as a preprocessing technique did not improve the deep learning models but helped the classical models to attain better recall sometimes. We also found that `Early Fusion' in general led to better performance over multiple problem setup than \emph{Late Fusion}. This indicates the importance of learning non-trivial interaction components between domain knowledge and learned representations from longitudinal medical history of the patients. 
It should be noted that our modeling scenario is more restrictive in the sense that there is no overlap of patients between the different folds (such as in the training and in the test data sets). 
\begin{table}[th!]
  \centering
  \caption{Global Feature importance}
  \label{tab:feat_imp_all}
    \scriptsize

  \begin{subtable}[t]{0.548\textwidth}
\centering
      \caption{Readmission Prediction}
      \label{tab:feat_imp:readmission}
    
      \begin{tabular}{l>{\raggedright\arraybackslash}p{0.48\linewidth}r}
\toprule
 Cateogry &                                       CCS CATEGORY &  Importance  \\
\midrule
     PROC &                                       Hemodialysis &  2.80549 \\
     ICD9 &                        Deficiency and other anemia &  1.98106 \\
     PROC &                                  Blood transfusion &  1.66835 \\
     PROC &                                  Arthroplasty knee & -1.54134 \\
     ICD9 &                      Acute cerebrovascular disease & -1.52745 \\
     ICD9 &                             Secondary malignancies & -1.48642 \\
     ICD9 &                        Transient cerebral ischemia & -1.48194 \\
     ICD9 &  Aortic; peripheral; and visceral artery aneurysms & -1.34200 \\
     ICD9 &      Other and ill-defined cerebrovascular disease & -1.21932 \\
     PROC &  Insertion; revision; replacement; removal of cardiac pacemaker or cardioverter/defibrillator & -1.18553 \\
     PROC &            Endarterectomy; vessel of head and neck & -1.13404 \\
     PROC &       Computerized axial tomography (CT) scan head &  1.11995 \\
     ICD9 &                                     Osteoarthritis & -1.11774 \\
     ICD9 &                Other non-epithelial cancer of skin & -1.06222 \\
   {\it Domain} &   {\it Charlson index}  &  1.05767 \\
     PROC &          Laminectomy; excision intervertebral disc &  1.05620 \\
     ICD9 &                Other ear and sense organ disorders & -1.04940 \\
     PROC &  Diagnostic cardiac catheterization; coronary arteriography & -1.03238 \\
     PROC &                  Electrographic cardiac monitoring &  1.02820 \\
     ICD9 &                                Malaise and fatigue & -1.01513 \\
\bottomrule
\end{tabular}
   \end{subtable}
  \begin{subtable}[t]{0.448\textwidth}
\centering
      \caption{Mortality Prediction}
      \label{tab:feat_imp:mortality}
    
      \begin{tabular}{l>{\raggedright\arraybackslash}p{0.48\linewidth}r}
\toprule
 Category &                                       CCS Category &  Importance \\
\midrule
     ICD9 &  Delirium, dementia, and amnestic and other cognitive &    0.046376 \\
     ICD9 &          Congestive heart failure; nonhypertensive &    0.040182 \\
     ICD9 &  Respiratory failure; insufficiency; arrest  &    0.034240 \\
     ICD9 &                       Residual codes; unclassified &    0.033168 \\
     Domain &                                     Charlson Index &    0.031612 \\
     ICD9 &                Acute and unspecified renal failure &    0.030669 \\
     ICD9 &                        Deficiency and other anemia &    0.027499 \\
     ICD9 &                               Cardiac dysrhythmias &    0.027440 \\
     ICD9 &                      Disorders of lipid metabolism &    0.026994 \\
     ICD9 &    Pneumonia (except that caused by tuberculosis)  &    0.026305 \\
     ICD9 &                             Essential hypertension &    0.025222 \\
     {\it Domain}      & {\it Length of stay} &    0.022936 \\
     ICD9 &                           Nutritional deficiencies &    0.022277 \\
     ICD9 &                           Urinary tract infections &    0.021677 \\
     ICD9 &  Chronic obstructive pulmonary disease and bronchiectasis &    0.021063 \\
     ICD9 &                   Other gastrointestinal disorders &    0.018653 \\
     ICD9 &                                    Other aftercare &    0.017234 \\
     ICD9 &   Coronary atherosclerosis and other heart disease &    0.015617 \\
     ICD9 &                       Septicemia (except in labor) &    0.015360 \\
\bottomrule
\end{tabular}

   \end{subtable}
\end{table}
 
\textbf{What are the key drivers - is knowledge important?}
We evaluated the importance of knowledge by ascertaining 
their importance with respect to raw data features at a model level. 
However, due to the black-box nature of deep models, obtaining global importance is a non-trivial task. 
We obtained this in a post-hoc manner by conducting a 
study to identify globally important features using a
logistic regression model (self-interpretable) as an explainer model.
Tables~\ref{tab:feat_imp:readmission}  and ~\ref{tab:feat_imp:mortality} 
show the most important features identified at a global
level for 30-day unplanned readmission prediction and 30 
day unexpected mortality, respectively. It can be seen that many 
data-driven features, spanning both diagnosis and 
procedure codes, are identified as important. This highlights how critical it is to model the longitudinal medical history for a patient. However, we also found that domain knowledge features, such as \emph{Charlson comorbidity} (for both readmission and mortality) and \emph{Length of stay} (for mortality), are important measures. This highlights the importance of blending domain knowledge into our deep models.

{\bf Where does the model perform well - Personalized Model
Performance}
Performance of clinical risk scores is typically reported at an aggregate level for an entire study population. 
However, patients in the population may vary
significantly from one another (e.g., patients with  vs. without
cardiovascular disease; males vs. female; patients in different age ranges.
In this respect, it is critical to characterize the
performance of the risk score across patient
sub-populations, so that clinicians can understand when a risk score is expected to be the most applicable to an individual patient.
More specifically, knowing the operating ranges of a model can lead to more confident use of AI models. 
To support this need, 
we characterized the performance across predefined sub-cohorts of patients. In a post hoc analysis on a hold-out evaluation test set that was scored with the trained risk prediction model, we grouped patients based on a
variety of 
baseline characteristics and recomputed prediction
performance measures on the patient cohort.

\begin{figure}[h]
  \centering
  \begin{subfigure}{.85\textwidth}
    \centering
    {\includegraphics[width=\linewidth]{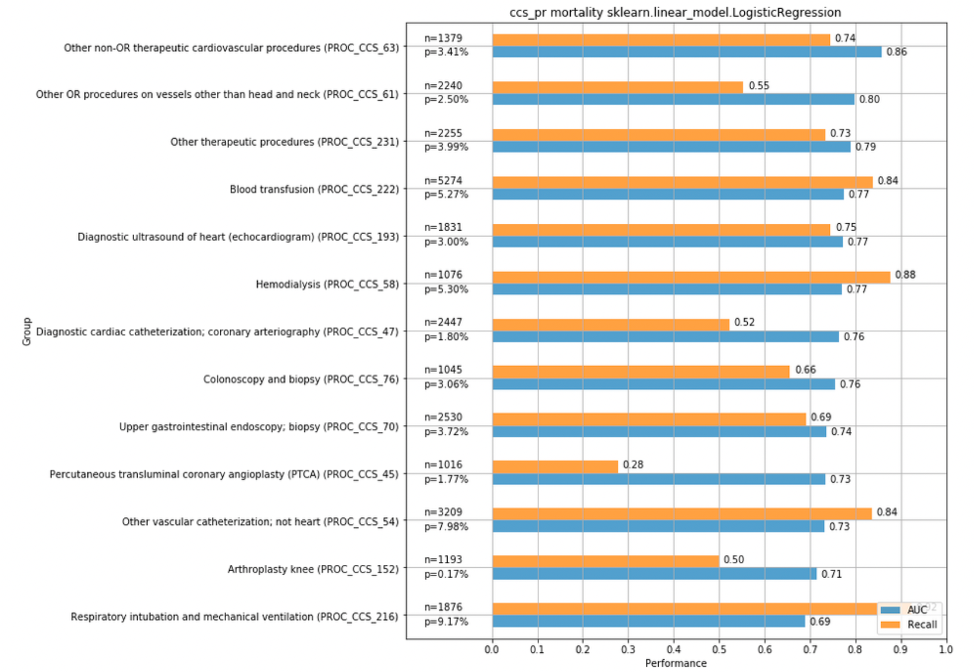}}
    \caption{CCS procedure categories\label{fig:ccs}
}
  \end{subfigure}
  \\
  \vspace{2em}
  \begin{subfigure}{.47\textwidth}
    \centering
    {\includegraphics[width=0.82\linewidth]{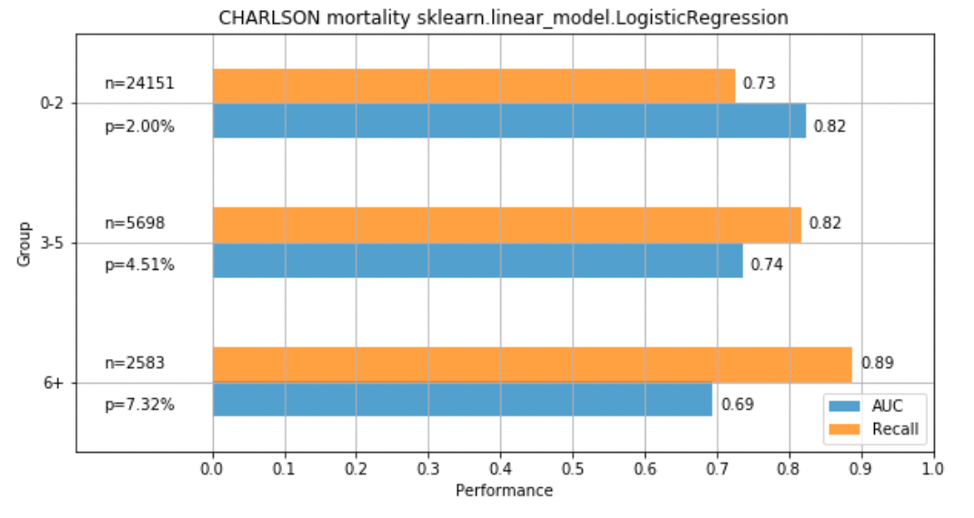}}
    \caption{Charlson index\label{fig:cci}}
   \end{subfigure}
\begin{subfigure}{.47\textwidth}
    {\includegraphics[width=0.92\linewidth]{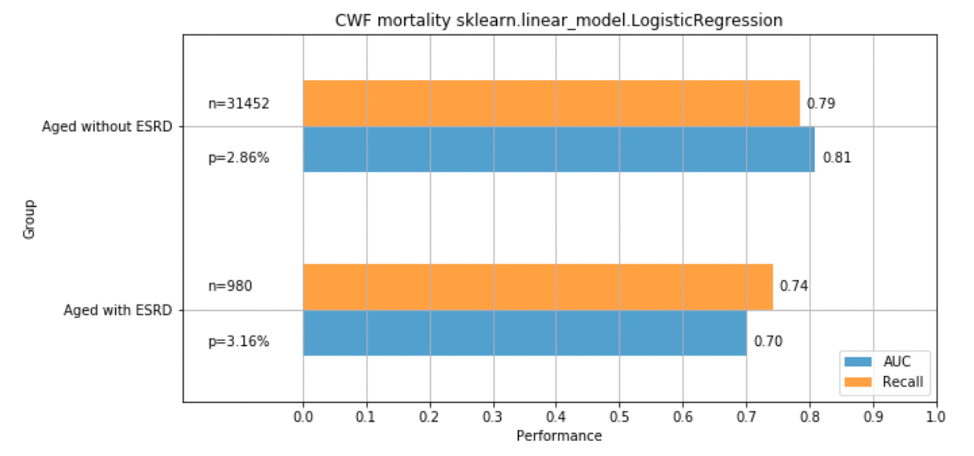}}
    \caption{CWF BENE MDCR status\label{fig:mdcr}}
  \end{subfigure}

  \caption{Performance (AUROC and Recall) for patient subgroups based on (a) CCS procedures, (b) Charlson Index, and (c) medicare 
  enrollment reasons. We see variations of model performance across sub-categories in (a-c).
  \label{fig:subgroup}}
\end{figure}

The 
investigated subgroups consisted of age, gender, race,
CCS procedure categories, and Charlson index. We report $3$ of the most interesting analysis in Figure~\ref{fig:subgroup}.
Figure~\ref{fig:ccs} shows model performance for population subgroups based on CCS procedure categories, for 30 day mortality prediction. There is variability in performance across groups. The performance is significantly higher than the population average (AUC = 0.80) for patients with \emph{Other non-OR therapeutic cardiovascular procedures (CCS 63)} (AUC = 0.86). It is much worse for patients that had \emph{Percutaneous transluminal coronary angioplasty (PTCA) (CCS 45)} (AUC = 0.73), \emph{Other vascular catheterization; not heart (CCS 54)} (AUC = 0.73), \emph{Arthroplasty knee (CCS 152)} (AUC = 0.71), and \emph{Respiratory intubation and mechanical ventilation (CCS 216)} (AUC = 0.69). This indicates that, while the model performs well at a population level, for patients with history of procedures (e.g, \emph{Arithroplasty of Knee}), the model should be used with caution. Figure~\ref{fig:cci} illustrates a non-trivial variability in performance for patient subgroups based on their Charlson index.
Patient groups with an index between 0-2 exhibit a slightly better performance (AUC = 0.82) than the population average (AUC = 0.80), while patient groups with higher indices, 3-5 and 6+ exhibit worse performance (AUC = 0.74 and 0.69, respectively).
This indicates that at these Charlson Index ranges, the model is less certain. Figure~\ref{fig:mdcr} depicts the performance for patient subgroups based on their \emph{CWF BENE MDCR} status. Almost all patients are in the “Aged without ESRD” group so the mortality prevalence (2.9\%) and model performance (AUC = 0.81) are roughly the same as the population average (p = 2.9\%, AUC = 0.80).  Only 980 patients are in the \emph{Aged with ESRD} group, which exhibits a higher mortality prevalence (3.2\%) but substantially worse performance (AUC = 0.70).

\section*{Conclusion}
  Deep learning models can suffer in performance when data is highly sparse and irregular, as is commonly the case for observational patient data. When sufficient domain knowledge is available, classical models using hand-crafted features can supplement deep models. We proposed a novel modeling framework that  
enjoy the best of both worlds, i.e., the benefits of non-parametric and flexible sequential modeling of patient history, via deep learning, while receiving guidance from domain knowledge.
Our empirical investigation, using large insurance claims data illustrates that such a method works efficiently across two tasks at discharge time: unplanned readmission and unexpected mortality. We also analyzed our model performance to determine conditions when (or when not) such models are confident in their predictions. Possible extensions of this work may include applying this approach on other data modalities (e.g., EHR data) to enable the assessment of readmission and risks prior to discharge.

\makeatletter
\renewcommand{\@biblabel}[1]{\hfill #1.}
\makeatother

\end{document}